\documentclass[runningheads]{llncs}

\usepackage{eccv}

\usepackage{eccvabbrv}

\usepackage{graphicx}
\usepackage{booktabs}

\usepackage{pifont}
\usepackage[table,dvipsnames]{xcolor}

\definecolor{bblue}{RGB}{0,30,95}
\definecolor{rred}{RGB}{190,0,0}
\definecolor{mygray}{gray}{0.1}
\definecolor{ggray}{RGB}{127,127,127}
\definecolor{rowgray}{RGB}{245,245,245}

\usepackage{subcaption}
\usepackage{booktabs} % for professional tables
\usepackage{multirow}

\usepackage{mathtools}
\usepackage{algorithm}

\usepackage{algorithmicx}

\usepackage{graphicx}
\usepackage{adjustbox}
\usepackage{subcaption}
\usepackage{algpseudocode}
\usepackage{soul}
\usepackage{wrapfig}
\usepackage{breqn}

\newcommand{\conf}[1]{{\color{gray}{\tiny{[{#1}]\!}}}}
\newcommand{\cmark}{\checkmark}
\newcommand{\xmark}{\(\times\)}
\DeclareMathOperator{\stopgrad}{stopgrad}

\usepackage{amsmath}
\usepackage{amssymb}
\usepackage{mathtools}
\usepackage[hidelinks]{hyperref}

\usepackage[capitalize,noabbrev]{cleveref}

\usepackage{orcidlink}

\begin{document}

% ---------------------------------------------------------------
\title{Geometry-Anchored Transport Framework for Exemplar-Free Class-Incremental Learning} 

\titlerunning{Geometry-Anchored Transport for EFCIL}

\author{ Hongye Xu\inst{1}\orcidlink{0009-0007-1590-0211} \and Bartosz Krawczyk\inst{1}\orcidlink{0000-0002-9774-0106} }

\authorrunning{H.~Xu and B.~Krawczyk}

\institute{ Chester F. Carlson Center for Imaging Science, Rochester Institute of Technology, Rochester, NY 14623, USA\\ \email{\{hx5239,bartosz.krawczyk\}@rit.edu} }

\maketitle

\begin{abstract}
Exemplar-free class-incremental learning (EFCIL) requires stable decision boundaries within a shifting feature space. While maintaining class-conditional Gaussian statistics provides a principled classification strategy, these parametric summaries remain sensitive to \emph{anisotropic representation drift}. Existing methods often transport these statistics across tasks using a decoupled, post-hoc paradigm: optimizing a backbone without explicit geometric constraints can distort the legacy manifold, limiting the precision of retroactive alignment. In this paper, we formulate feature transport as an endogenous training constraint rather than a separate post-task step, presenting the \textbf{Geometry-Anchored Transport Framework}. First, we derive an \emph{Analytic Geometric Anchor} via Mahalanobis-aligned regression to mitigate macroscopic anisotropic drift. Second, we introduce a \emph{Topology-Aware Evolution} objective that regularizes localized manifold degradation while calibrating a residual network against the analytic prior. By coupling manifold evolution with transport constraints during the primary training phase, our framework mitigates evaluation errors without requiring decoupled fine-tuning. Experiments across CIFAR-100, TinyImageNet, and ImageNet-100 demonstrate that the proposed framework consistently improves upon existing post-hoc alternatives under strict exemplar-free constraints. The code is available at \url{https://github.com/HXuSz11/GATF_ECCV2026}.
\end{abstract}    
\section{Introduction}
\label{intro}

The primary challenge in Exemplar-Free Class-Incremental Learning (EFCIL) is managing the plasticity-stability trade-off within a non-stationary feature space~\cite{Parisi:2019,Lange:2022,zenke2017continual,hadsell2020embracing,kim2023stap}. Without the safety net of memory buffers~\cite{Boschini:2023}, models are vulnerable to catastrophic forgetting \cite{HacohenT25,mccloskey1989catastrophic,serra2018overcoming,chaudhry2018riemannian}. To address this, recent methods have increasingly adopted prototype-based and Gaussian-parametric inference~\cite{toldo2022fecam,petit2023fetril,rypesc2024task,GoswamiSLKT024,Magistri2024EFC}. By summarizing previously seen distributions into compact sufficient statistics~\cite{LyuWZSSZJ23}, networks can utilize Bayes or Mahalanobis decision rules across the sequence of learned tasks~\cite{toldo2022fecam,rypesc2024task}.

However, this parametric approach encounters the issue of \emph{representation drift}~\cite{KanebakoT21,toldo2022fecam,zhuang2022acil,petit2023fetril,0004TLHWCJ020,zhuang2024ds,rypesc2024task,Magistri2024EFC,KimKS25}. As the backbone assimilates novel categories, the underlying embedding space undergoes continuous non-linear deformation. To update prior statistics, the standard paradigm employs a decoupled "train-then-adapt" sequence~\cite{GoswamiSLKT024,GomezVillaGWBTW24,Magistri2024EFC,He_ICML2025_DPCR}. In this formulation, the backbone is first updated on new data using knowledge distillation, and subsequently, a post-hoc adapter is trained to map legacy prototypes into the modified space. 

This decoupled approach introduces two structural limitations. First, optimizing the backbone—driven primarily by the cross-entropy of new classes and standard distillation—can induce \emph{topological degradation}~\cite{douillard2020podnet,hadsell2020embracing,kim2023stap}. Evolving the network without cross-space geometric constraints can distort the neighborhood structure of prior categories, complicating subsequent post-hoc mappings. Second, Mahalanobis-based evaluation is sensitive to \emph{anisotropic drift}. Transport errors along low-variance directions are amplified by the inverse covariance matrix, which can erode decision margins and introduce classification bias between old and new classes.

\begin{figure}[t] 
  \centering
  
  % ================= Baseline & CIFAR-100 =================
  \begin{subfigure}[b]{0.37\linewidth}
    \centering
    \includegraphics[width=\linewidth]{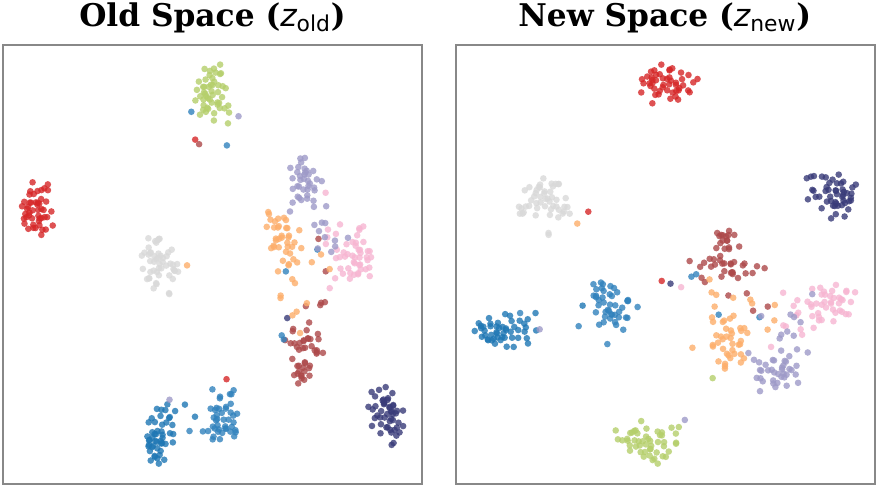}
    \caption{Baseline (Task 0)}
  \end{subfigure}
  \hfill
  \begin{subfigure}[b]{0.37\linewidth}
    \centering
    \includegraphics[width=\linewidth]{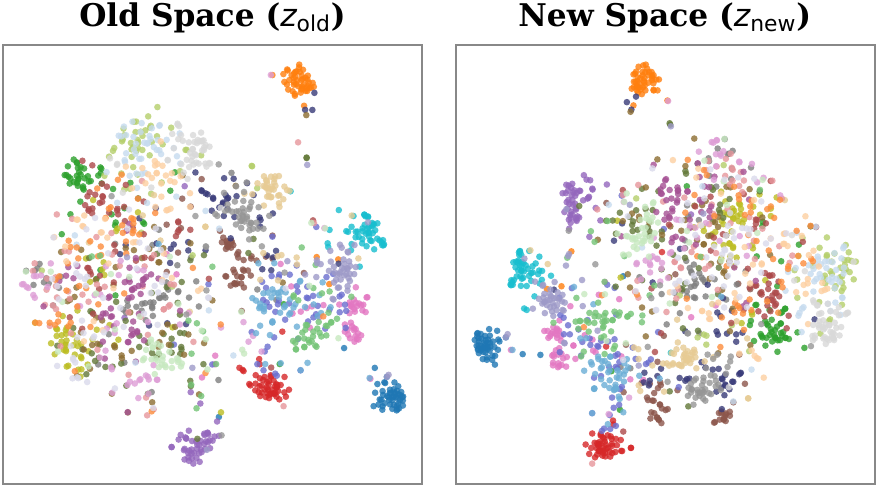}
    \caption{Baseline (Task 2)}
  \end{subfigure}
  \hfill
  \vrule width 0.8pt 
  \hfill
  \begin{subfigure}[b]{0.205\linewidth} 
    \centering
    \includegraphics[width=\linewidth]{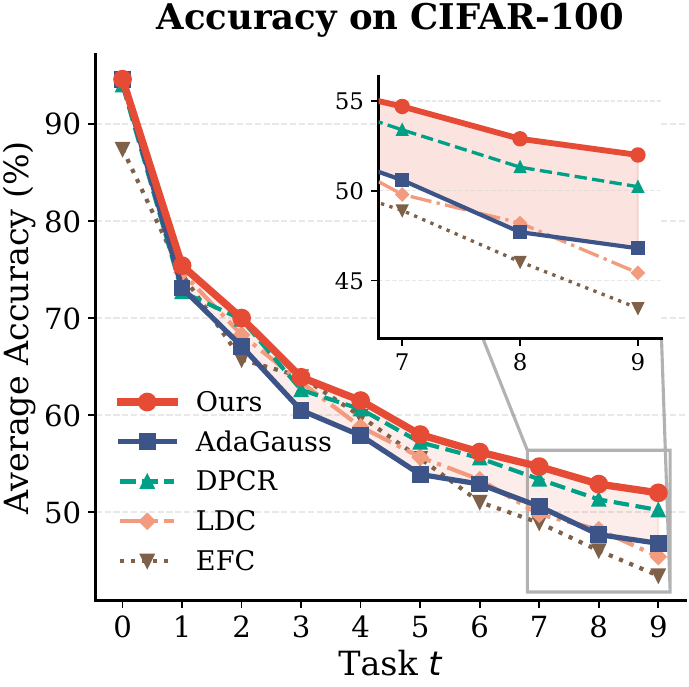}
    \caption{CIFAR-100}
  \end{subfigure}
  
  \vspace{3mm} 
  
  % ================= Ours & TinyImageNet =================
  \begin{subfigure}[b]{0.37\linewidth}
    \centering
    \includegraphics[width=\linewidth]{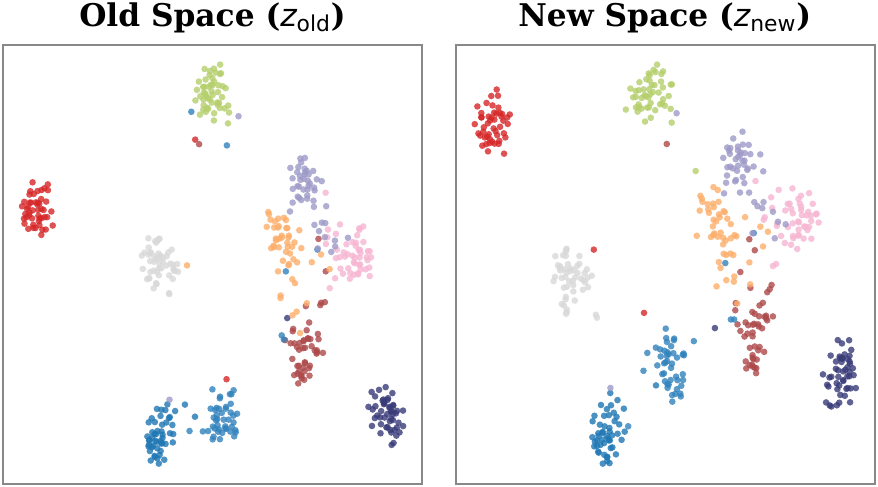}
    \caption{Ours (Task 0)}
  \end{subfigure}
  \hfill
  \begin{subfigure}[b]{0.37\linewidth}
    \centering
    \includegraphics[width=\linewidth]{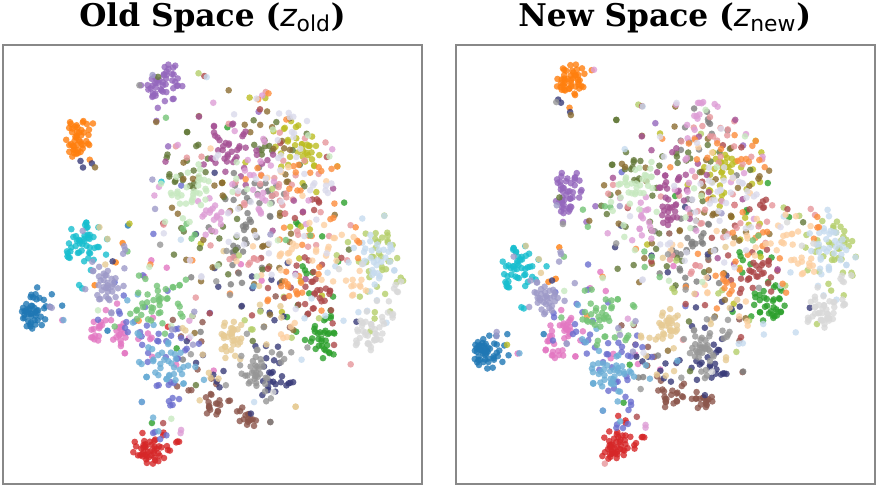}
    \caption{Ours (Task 2)}
  \end{subfigure}
  \hfill
  \vrule width 0.8pt
  \hfill
  \begin{subfigure}[b]{0.205\linewidth}
    \centering
    \includegraphics[width=\linewidth]{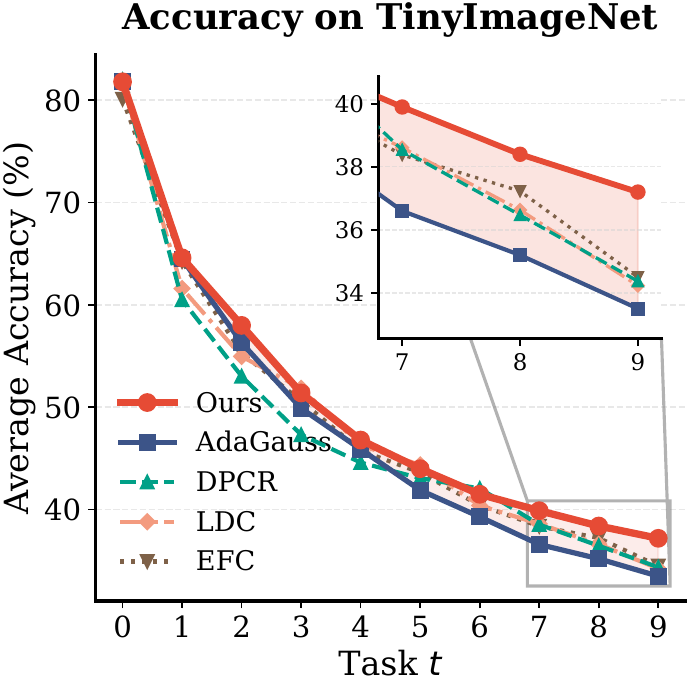}
    \caption{TinyImageNet}
  \end{subfigure}
  
  \vspace{-1mm}
  \caption{\textbf{Topological anchoring and continual learning performance.} 
    \textbf{(Left \& Middle):} Feature transport from the frozen model ($z_{\mathrm{old}}$) to the new model ($z_{\mathrm{new}}$) visualized on TinyImageNet. In the baseline (AdaGauss~\cite{rypesc2024task}) (\textbf{Top}), the embedding space undergoes significant rotation and non-uniform displacement, leading to a disruption of the relative spatial relationships between categories. Conversely, our framework (\textbf{Bottom}) anchors the global topology, effectively mitigating these geometric distortions to maintain structural alignment across tasks. 
    \textbf{(Right):} This spatial preservation corresponds to sustained accuracy, improving upon established baselines. A detailed discussion regarding our baseline selection is provided in the \textsc{Supplementary}.}
  \label{fig:teaser_2x3}
  \vspace{-3mm}
\end{figure}

To address these limitations, we propose the \textbf{Geometry-Anchored Transport Framework}, integrating feature transport as a geometry-aware constraint during representation learning rather than treating it solely as a post-hoc adjustment. 

The framework relies on an \textbf{Analytic Geometric Anchor (AGA)}. Using a closed-form generalized least squares (GLS) prior aligned with the Mahalanobis geometry, we establish a baseline mapping to mitigate macroscopic anisotropic drift and suppress tail-end errors. However, applying a static analytic prior to an evolving neural network necessitates topological stability. Consequently, we structure the backbone's training through \emph{Topology-Aware Evolution}. By incorporating a local anchoring constraint directly into the optimization trajectory, we restrict the backbone from parameter updates that degrade legacy neighborhood structures. This encourages smoother backbone evolution, allowing a residual network to calibrate against the non-linear deformations not captured by the analytic prior, thereby reducing the reliance on decoupled post-hoc fine-tuning.

\noindent\textbf{Contributions.} Our contributions are summarized as follows:
\begin{itemize}
  \item We analyze the limitations of the decoupled post-hoc transport paradigm, showing that topological degradation and Mahalanobis-amplified anisotropic drift contribute to margin erosion in EFCIL.
  \item We introduce an Analytic Geometric Anchor formulated via a Sylvester equation, designed to mitigate macroscopic geometric drift and bound low-variance transport errors.
  \item We propose a topology-aware evolution strategy that calibrates a non-linear residual against the analytic prior while grounding the active feature space. This coupled framework provides margin stability bounds (Theorem~2) and demonstrates competitive performance across CIFAR-100, TinyImageNet, ImageNet-100, and CUB-200.
\end{itemize}

\begin{figure*}[t]
  \centering
  % Make sure to update the image file to visually match the new "Geometry-Anchored" block diagram later.
  \includegraphics[width=\textwidth]{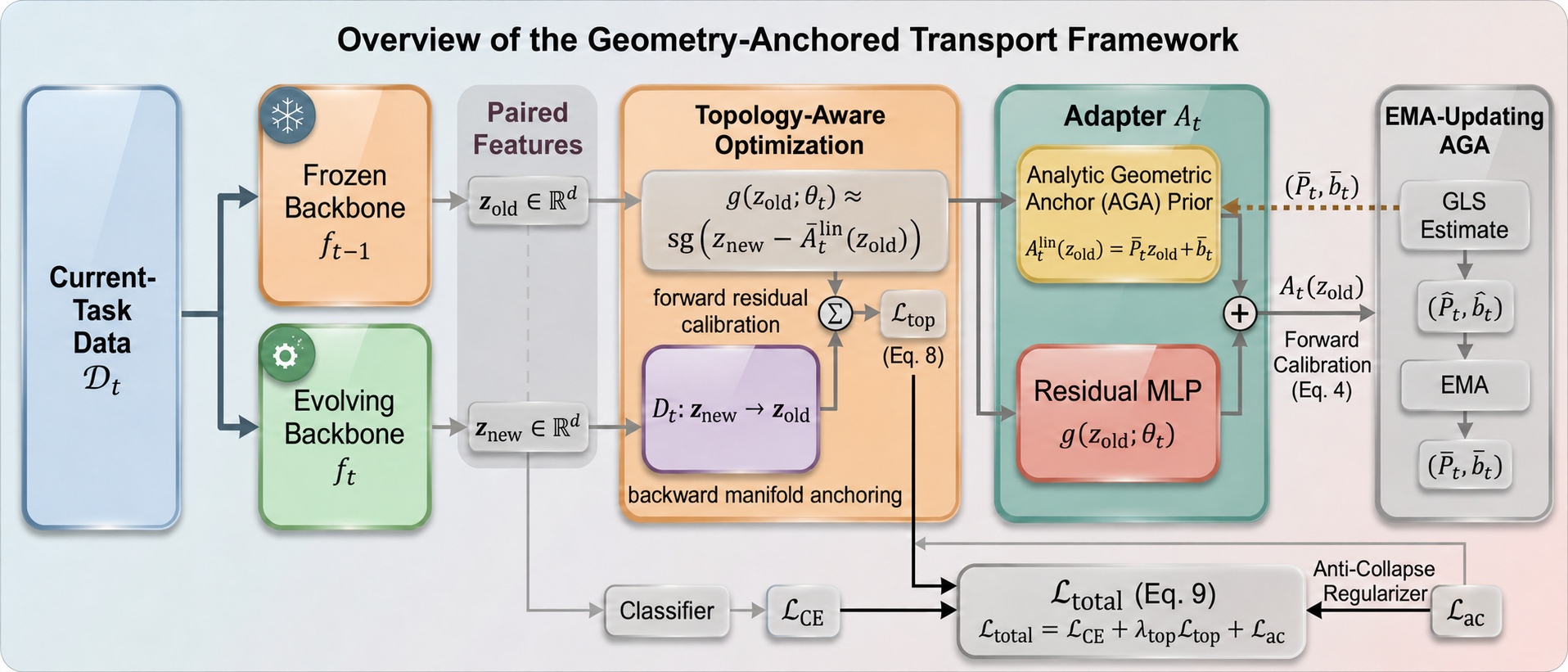}
  \vspace{-6mm}
  \caption{\textbf{Overview of the Geometry-Anchored Transport Framework.} At task $t$, the active backbone $f_t$ learns from the new dataset $\mathcal{D}_t$, while the frozen $f_{t-1}$ provides legacy features $z_{\text{old}}$. Rather than relying solely on decoupled post-hoc adaptation, we integrate feature transport into the training phase. \textbf{Macroscopically}, an Analytic Geometric Anchor (AGA) computes a closed-form linear prior to mitigate anisotropic drift. \textbf{Microscopically}, the backbone utilizes Topology-Aware Evolution: a backward distiller $D_t$ anchors the legacy manifold to restrict topological degradation, and a forward residual network $g$ calibrates against the AGA's displacement. Ultimately, this coupled optimization yields a geometry-aware adapter $A_t$ that facilitates the stable transport of legacy Gaussian statistics for robust Mahalanobis evaluation.}
  \label{fig:main_pipline}
  \vspace{-2mm}
\end{figure*}
\section{Related Work}
\label{related}

\paragraph{\textbf{Geometry-Aware Continual Evaluation.}}
To mitigate catastrophic forgetting without storing raw samples, recent research has focused on adapting the evaluation geometry. Since standard Euclidean metrics may degrade under representation drift, approaches such as FeCAM~\cite{toldo2022fecam} utilize anisotropic, class-conditional covariances to construct geometry-aware decision boundaries. Similarly, FeTrIL~\cite{petit2023fetril} freezes the feature extractor early and synthesizes pseudo-features to maintain old-class separability. While these methods improve stability, their reliance on largely static feature spaces can restrict plasticity for new distributions. Our framework utilizes Mahalanobis evaluation but avoids the static-backbone constraint. Instead, we regularize the evaluation geometry during continuous training via an explicit analytic transport anchor.

\paragraph{\textbf{Predictive Feature Mapping and Transport.}}
To maintain plasticity while updating prior statistics, recent EFCIL methods map legacy representations into the active feature space. Initial approaches, such as SDC~\cite{0004TLHWCJ020} and ADC~\cite{GoswamiSLKT024}, estimated global or class-wise translations. Subsequent methods, including LDC~\cite{GomezVillaGWBTW24}, EFC~\cite{Magistri2024EFC}, AdaGauss~\cite{rypesc2024task}, and DPCR~\cite{He_ICML2025_DPCR}, employ auxiliary neural networks to project cached prototypes forward based on pairwise feature relationships. 

Despite their strong empirical performance, these architectures often rely on a decoupled, post-hoc paradigm. The active backbone is optimized largely independently of the downstream transport mechanism, typically constrained by standard distillation. This separation implies that the post-hoc adapter must map features across a manifold that may undergo topological degradation during the primary optimization phase. Furthermore, auxiliary neural adapters typically optimize isotropic Euclidean objectives, which may under-penalize the low-variance anisotropic distortions that affect Bayes classifiers. Our methodology addresses this decoupled structure. By solving macroscopic drift analytically and embedding a topology-aware constraint directly into the backbone optimization, we constrain manifold evolution to bound transport errors during the primary training phase.

\paragraph{\textbf{Relation to DPCR.}}
DPCR~\cite{He_ICML2025_DPCR} is the closest recent post-hoc alternative to our framework. It estimates
semantic shift after backbone training through dual projection and reconstructs
a classifier with ridge regression. In contrast, our method does not aim to
repair the classifier after the representation has already changed. Instead, the
AGA provides a Mahalanobis-aligned analytic prior during backbone optimization,
and the topology-aware objective constrains feature evolution before old-class
geometry is distorted. Thus, the central difference is not merely the form of the
transport map, but the timing of the constraint: DPCR performs post-training
calibration, whereas our framework enforces training-time geometry anchoring
to make the final statistic pushforward more reliable.
\section{Methodology}
\label{Method}

\subsection{Setup and Evaluation Space}

We study an exemplar-free class-incremental learning problem with tasks indexed
by $t\in\{0,1,\dots\}$. After completing task $t{-}1$, we freeze the previous
backbone $f_{t-1}$ and train a new backbone $f_t$ using only the current-task
dataset $\mathcal{D}_t$. For an input $x$, we denote:
\begin{equation}
z_{\mathrm{old}}=f_{t-1}(x)\in\mathbb{R}^d,
\qquad
z_{\mathrm{new}}=f_t(x)\in\mathbb{R}^d.
\end{equation}

Evaluation is performed in the active feature space induced by $f_t$ using
class-conditional Gaussian statistics. For newly introduced classes
$c\in\mathcal{C}_t$, the moments $(\mu_c^t,\Sigma_c^t)$ are estimated directly
from $\mathcal{D}_t$ under $f_t$. For old classes
$c\in\mathcal{C}_{1:t-1}$, the archived statistics
$(\mu_c^{t-1},\Sigma_c^{t-1})$ remain untouched during task-$t$ training.
After training converges, we perform a training-free distribution pushforward:
\begin{equation}
\tilde z_m\sim \mathcal{N}(\mu_c^{t-1},\Sigma_c^{t-1}),\qquad
v_m=A_t(\tilde z_m),
\end{equation}
and re-estimate $(\hat\mu_c^t,\hat\Sigma_c^t)$ from
$\{v_m\}_{m=1}^M$ in the active feature space. We denote the assembled
task-$t$ statistics for all seen classes by $(\mu_c^t,\Sigma_c^t)$, where old
classes use the transported estimates and new classes use empirical estimates.

Unless stated otherwise, inference uses a Gaussian Bayes/Mahalanobis rule over
all seen classes:
\begin{equation}
s_c(x)=
(z-\mu_c^t)^\top(\Sigma_c^t)^{-1}(z-\mu_c^t),
\qquad
\hat y=\arg\min_{c\in\mathcal{C}_{1:t}} s_c(x),
\end{equation}
where $z=f_t(x)$.

To relate $z_{\mathrm{old}}$ and $z_{\mathrm{new}}$, we incorporate a
co-evolving adapter $A_t$. Unlike prior methods that relegate transport to a
post-hoc stage, we treat $A_t$ as an integral constraint during primary training,
so that the feature space evolves in a form that remains compatible with the
task-end pushforward of old-class Gaussian statistics.

\subsection{Geometry-Anchored Transport Framework}

Bridging two independently evolving high-dimensional spaces presents distinct optimization challenges. A conventional approach decouples feature learning from cross-space alignment, optimizing the backbone independently of legacy structures before fitting a post-hoc adapter. However, this decoupled paradigm can limit overall stability: if stable classification relies on bounded transport errors (Theorem~\ref{theorem2}), the backbone's evolutionary trajectory should be explicitly constrained to preserve these bounds. Unconstrained optimization leaves the mapping vulnerable to anisotropic representation drift—where distortions along low-variance yet decision-critical directions can shift Mahalanobis margins. 

To address this, our framework approaches the transport problem by introducing an explicit Analytic Geometric Anchor (AGA) to resolve macroscopic anisotropic drift analytically. This is supported by a Topology-Aware Evolution mechanism that regularizes the backbone's local deformations during the primary training phase.

\subsubsection{The Analytic Geometric Anchor (AGA).}
To make old$\!\to$new transport geometrically aligned, we anchor the forward adapter $A_t$ with a closed-form linear prior, which we term the Analytic Geometric Anchor (AGA). We parameterize the forward transport as an explicitly decomposed function:
\begin{equation}
\label{eq:res_adapter}
A_t(z)
=
A^{\mathrm{lin}}_t(z)+g(z;\theta_t),
\qquad
A^{\mathrm{lin}}_t(z)=P_t z + b_t,
\end{equation}
where $P_t\in\mathbb{R}^{d\times d}$ and $b_t\in\mathbb{R}^{d}$ define an affine map representing the AGA, and $g(\cdot;\theta_t)$ is a small MLP learning localized residual corrections. 

The affine component captures the dominant global drift, while the residual
network models localized non-linear corrections. During training, the linear
component is used as a zero-gradient prior: gradients are not propagated through
the closed-form estimate, but the active backbone and residual network are
optimized against it.

We estimate this linear relationship using paired current-task features
$x\sim\mathcal{D}_t$, with
$\mu_{\mathrm{old}}=\mathbb{E}[z_{\mathrm{old}}]$,
$\mu_{\mathrm{new}}=\mathbb{E}[z_{\mathrm{new}}]$, and
$C_{\mathrm{on}}=\mathrm{Cov}(z_{\mathrm{old}},z_{\mathrm{new}})$.
To target Mahalanobis-sensitive tail errors (empirically analyzed in Sec.~4.2),
we formulate the estimation of $P_t,b_t$ as a ridge-regularized Generalized
Least Squares (GLS) problem:
\begin{equation}
\label{eq:gls_obj}
\min_{P_t,b_t}\;
\mathbb{E}\Big[
\big\|(P_t z_{\mathrm{old}}+b_t)-z_{\mathrm{new}}\big\|^2_{\Sigma_{\mathrm{new}}^{-1}}
\Big]
+\rho\|P_t\|_F^2.
\end{equation}
Centering yields the optimal intercept $b_t=\mu_{\mathrm{new}}-P_t\mu_{\mathrm{old}}$. The optimal mapping $P_t$ for our AGA admits a Sylvester-form~\cite{WuFDW10a} solution (Theorem~\ref{theorem1}), which penalizes transport mismatch along low-variance directions while providing anisotropic shrinkage to limit noise amplification. 

\begin{theorem}[AGA formulation via Ridge-GLS and anisotropic shrinkage]
\label{theorem1}
Let $z_{\mathrm{old}}, z_{\mathrm{new}}\in\mathbb{R}^d$ denote paired features,
and define centered variables $x=z_{\mathrm{old}}-\mu_{\mathrm{old}}$, $y=z_{\mathrm{new}}-\mu_{\mathrm{new}}$ with
$\Sigma_{\mathrm{old}}=\mathrm{Cov}(x)\succ 0$, $\Sigma_{\mathrm{new}}=\mathrm{Cov}(y)\succ 0$,
and $C_{\mathrm{on}}=\mathrm{Cov}(x,y)$.
For $\rho>0$, the ridge-regularized Mahalanobis regression (Eq.~\ref{eq:gls_obj}) leads to the unique minimizer satisfying $b^\star=\mu_{\mathrm{new}}-P^\star\mu_{\mathrm{old}}$, where
$P^\star$ is the unique solution of the Sylvester equation:
\begin{equation}
P^\star\Sigma_{\mathrm{old}}+\rho\,\Sigma_{\mathrm{new}}P^\star=C_{\mathrm{on}}^\top.
\end{equation}
Moreover, letting $\Sigma_{\mathrm{new}}=U\Lambda U^\top$ and $\Sigma_{\mathrm{old}}=V\Gamma V^\top$
with $\Lambda=\mathrm{diag}(\lambda_i)$ and $\Gamma=\mathrm{diag}(\gamma_j)$, the solution admits the spectral form
$P^\star=UXV^\top$ where
\begin{equation}
X_{ij}=\frac{(U^\top C_{\mathrm{on}}^\top V)_{ij}}{\gamma_j+\rho\,\lambda_i}.
\end{equation}
\end{theorem}

\subsubsection{Topology-Aware Evolution and the Overall Objective.}
While the AGA aligns the macroscopic geometry, directly superimposing it on an unconstrained, rapidly evolving backbone can introduce instability. Open-loop optimization of $f_t$ naturally induces local topological degradation of the manifold, creating non-linear local distortions that the global linear prior $\bar P_t$ cannot capture. 

To bridge this gap, we integrate a Topology-Aware Evolution objective directly
into the active training phase. During training, the AGA prior used in this
objective is the EMA-smoothed affine map
$\bar A_t^{\mathrm{lin}}(z)=\bar P_tz+\bar b_t$. We define a joint topological
penalty $\mathcal{L}_{\mathrm{top}}$ that couples the forward residual network
$g$ and the backward distiller $D$:
\begin{equation}
\label{eq:ltop}
\mathcal{L}_{\mathrm{top}} = 
\underbrace{\Big\|g(z_{\mathrm{old}};\theta_t) -
\stopgrad\Big(z_{\mathrm{new}} -
\overbrace{\bar A^{\mathrm{lin}}_t(z_{\mathrm{old}})}^{\textbf{AGA Prior}}
\Big)\Big\|_2^2}_{\text{forward residual calibration}}
\;+\;
\underbrace{\|D(z_{\mathrm{new}})-z_{\mathrm{old}}\|_2^2}_{\text{backward manifold anchoring}}.
\end{equation}
The overall end-to-end training objective for the active backbone and transport modules is thus formulated as:
\begin{equation}
\label{eq:ltotal}
\mathcal{L}_{\mathrm{total}}
=
\mathcal{L}_{\mathrm{CE}}
+ \lambda_{\mathrm{top}} \mathcal{L}_{\mathrm{top}}
+ \mathcal{L}_{\mathrm{ac}},
\end{equation}
where $\mathcal{L}_{\mathrm{ac}}$ denotes the anti-collapse loss proposed by AdaGauss~\cite{rypesc2024task}. This unified formulation explicitly binds the network's evolution to the geometric prior. In $\mathcal{L}_{\mathrm{top}}$, the forward residual calibration directs the shallow MLP $g(\cdot;\theta_t)$ to absorb the non-linear displacement that the analytic AGA does not capture. The $\stopgrad$ operator ensures that the residual tracks the evolving space without artificially limiting the backbone's plasticity for the new task. 

Simultaneously, the backward manifold anchoring restricts structural collapse. Instead of strictly penalizing Euclidean drift (which limits plasticity), we employ the auxiliary distiller $D$ to ensure the new representation retains sufficient geometric integrity to reconstruct the legacy manifold. Its gradients flow directly into $f_t$, penalizing the backbone for deformations that would disrupt legacy neighborhoods. Together, these joint constraints restrict the backbone to topology-preserving deformations, ensuring the representation respects the bounds required for stable Bayes evaluation (Theorem~\ref{theorem2}).

\subsection{EMA-Updating the AGA During Backbone Training}

Because $f_t$ is optimized dynamically via SGD, the joint distribution of $(z_{\mathrm{old}},z_{\mathrm{new}})$ evolves continuously. Instantaneous closed-form estimates of $(P_t,b_t)$ can exhibit high-variance fluctuations, which may destabilize the residual calibration objective. We therefore maintain an exponential moving average (EMA) of the linear prior.

\subsubsection{Periodic prior refresh \& EMA smoothing.}
Every $\Delta$ iterations, we recompute an instantaneous GLS estimate $(\widehat P_t^{(k)},\widehat b_t^{(k)})$ from a short stream of paired features. We update a running prior $(\bar P_t,\bar b_t)$ for the AGA with momentum $m\in[0,1)$:
\begin{equation}
\label{eq:ema_prior}
\bar P_t \leftarrow m\,\bar P_t+(1-m)\,\widehat P_t^{(k)},
\qquad
\bar b_t \leftarrow m\,\bar b_t+(1-m)\,\widehat b_t^{(k)}.
\end{equation}
Between refreshes, the adapter uses $\bar A^{\mathrm{lin}}_t(z)=\bar P_t z+\bar b_t$. By treating $(\bar P_t,\bar b_t)$ as constants during backpropagation, the EMA prior provides a smooth, lower-variance geometric baseline.

\subsection{Theoretical Guarantees for Bayesian Evaluation}

As defined in Sec.~3.1, evaluation relies on Mahalanobis scores computed from
class-conditional Gaussian statistics in the active feature space. The reliability
of this evaluation depends on how accurately old-class moments are transported
through $A_t$. We formalize this dependency by establishing a stability bound on
the decision margin:
\begin{theorem}[Margin stability under transported moment errors]
\label{theorem2}
After task $t$, let the Bayes/Mahalanobis score be $ d_c(z)=(z-\mu_c)^\top \Sigma_c^{-1}(z-\mu_c)$, with prediction $\hat y(z)=\arg\min_c d_c(z)$. Define the old-class margin for a labeled point $(z,y)$ as $m(z)=\min_{c\neq y} d_c(z)-d_y(z)$. Assume for each class $c$ we use transported parameters $(\hat\mu_c,\hat\Sigma_c)$ with perturbations $\Delta\mu_c=\hat\mu_c-\mu_c$ and $\Delta\Sigma_c=\hat\Sigma_c-\Sigma_c$ satisfying $\|\Delta\mu_c\|_2\le \eta_\mu$ and $\|\Delta\Sigma_c\|_2\le \eta_\Sigma$. Let $\lambda_{\min,c}$ denote the smallest eigenvalue of $\Sigma_c$ and assume $\eta_\Sigma < \tfrac{1}{2}\lambda_{\min,c}$ for all $c$.

Then for any feature $z$, the deviation of each class score is uniformly bounded by:
\begin{equation}
\begin{aligned}
\big|\hat d_c(z)-d_c(z)\big|
&\le
\underbrace{\frac{2}{\lambda_{\min,c}}\|z-\mu_c\|_2\,\eta_\mu+\frac{1}{\lambda_{\min,c}}\eta_\mu^2}_{\text{mean error term}} \\
&\quad+\;
\underbrace{\frac{2\,\eta_\Sigma}{\lambda_{\min,c}^2}\big(\|z-\mu_c\|_2+\eta_\mu\big)^2}_{\text{covariance error term}}.
\end{aligned}
\end{equation}
Consequently, letting $\varepsilon_c(z)$ denote the bound above and $\varepsilon(z)=\varepsilon_y(z)+\max_{c\neq y}\varepsilon_c(z)$, we guarantee:
\begin{equation}
m(z)>\varepsilon(z)\quad\Longrightarrow\quad \arg\min_c \hat d_c(z)=y,
\end{equation}
meaning the Bayes/Mahalanobis prediction remains identical despite the transport perturbations.
\end{theorem}

\subsubsection{Implications for our framework.}
Theorem~\ref{theorem2} connects robust Mahalanobis evaluation to the accuracy
of transported means and covariances. The AGA suppresses macroscopic
anisotropic drift through the Mahalanobis-aligned closed-form prior
(Theorem~\ref{theorem1}), while the topology-aware objective limits local
manifold degradation during backbone optimization. Together, these two
components reduce the perturbation terms $\eta_\mu$ and $\eta_\Sigma$ that
directly control the margin-stability condition.

After task training, the final adapter is instantiated as
\begin{equation}
A_t(z)=\bar P_tz+\bar b_t+g(z;\theta_t).
\end{equation}
The analytic term accounts for the dominant global transport, and the residual
term captures the remaining localized deformation. As defined in Sec.~3.1,
old Gaussian statistics are then pushed forward through $A_t$ in a
training-free stage. This design keeps transport coupled with representation
learning during optimization, while avoiding an additional post-hoc adapter
fine-tuning stage. Our ablation in Sec.~4.2 further verifies that the end-to-end
adapter is already effective after co-evolution, and that optional post-hoc
refinement is generally unnecessary.

\section{Experiments}
\label{sec:experiments}

\paragraph{\textbf{Evaluation setup and baselines.}}
Following standard exemplar-free protocols, we compare against regularizers (EWC~\cite{kirkpatrick2017ewc}, LwF~\cite{LiH16}), prototype approaches (SDC~\cite{0004TLHWCJ020}, PASS~\cite{zhu2021pass}, FeTrIL~\cite{petit2023fetril}, FeCAM~\cite{toldo2022fecam}), and recent methods (EFC~\cite{Magistri2024EFC}, ADC~\cite{GoswamiSLKT024}, LDC~\cite{GomezVillaGWBTW24}, AdaGauss~\cite{rypesc2024task}, DPCR~\cite{He_ICML2025_DPCR}).
Baselines are reproduced using PyCIL~\cite{zhou2023pycil}, FACIL~\cite{masana2022class}, OCL~\cite{MAI202228}, or their official codes.
We evaluate on CIFAR-100~\cite{krizhevsky2009learning}, TinyImageNet~\cite{le2015tiny}, ImageNet-100~\cite{deng2009imagenet} (from scratch) and CUB-200~\cite{wah2011caltech} (pretrained) over $T{\in}\{10, 20\}$ tasks, reporting $A_{\text{last}}$ and running average $A_{\text{inc}}$.

\paragraph{\textbf{Implementation.}}
Our codebase builds upon AdaGauss~\cite{rypesc2024task}, preserving its prototype/Gaussian bookkeeping and classifier to isolate performance improvements to the proposed transport mechanism.
Unless stated otherwise, we train a ResNet-18~\cite{he2016deep} with batch size 256, project 512-D features into a 64-D prototype space, and maintain class-conditional Gaussians with a Gaussian Bayes classifier.
Optimization schedules follow AdaGauss (SGD, 200 epochs for CIFAR-100/TinyImageNet/ImageNet-100; CUB-200 uses a pretrained backbone with a separate head learning rate). Detailed hyperparameter settings and sensitivity experiments can be found in the \textsc{Supplementary Material}. We additionally verify in the \textsc{Supplementary Material} that a sampling-based
linear classifier trained from the same transported Gaussians gives comparable
performance, indicating that the improvement comes from better transported
statistics rather than a specific readout.

\paragraph{\textbf{Reproduction note.}}
All reproduced baselines are evaluated under the same class orders and protocol.
During reproduction of covariance- or projection-based methods, we observed that
some public implementations can encounter rank-deficient or nearly singular
covariance/projection matrices, which may break SVD, inverse, pseudo-inverse, or
Gaussian construction routines. We therefore apply standard numerical safeguards
such as symmetrization, diagonal loading, eigenvalue clipping, and jitter when
needed. These safeguards preserve the original objectives and are applied only
to ensure stable execution. For CUB-200, DPCR did not complete reliably under
the pretrained protocol, so we leave these entries blank rather than reporting
selectively successful runs.

\begin{table}[t]
\caption{\textbf{CIFAR-100 \& TinyImageNet (from scratch).}
Average incremental ($A_{\text{inc}}$) and last-task average ($A_{\text{last}}$) accuracy (\%). Results are averaged over five runs (standard deviations are provided in \textsc{Supplementary Material}). Best in \textbf{bold}.}
\label{tab:main_results}
\centering
\small
\begin{tabular*}{\linewidth}{@{\extracolsep{\fill}}lcccccccc}
\toprule
\multirow{2}{*}{\textbf{Method}} &
\multicolumn{4}{c}{\textbf{CIFAR-100}} &
\multicolumn{4}{c}{\textbf{TinyImageNet}} \\
\cmidrule(lr){2-5}\cmidrule(lr){6-9}
& \multicolumn{2}{c}{$T{=}10$} & \multicolumn{2}{c}{$T{=}20$} &
  \multicolumn{2}{c}{$T{=}10$} & \multicolumn{2}{c}{$T{=}20$} \\
\cmidrule(lr){2-3}\cmidrule(lr){4-5}\cmidrule(lr){6-7}\cmidrule(lr){8-9}
& $A_{\text{last}}$ & $A_{\text{inc}}$ &
  $A_{\text{last}}$ & $A_{\text{inc}}$ &
  $A_{\text{last}}$ & $A_{\text{inc}}$ &
  $A_{\text{last}}$ & $A_{\text{inc}}$ \\
\midrule
EWC &
  30.9 & 50.4 & 17.0 & 34.2 &
  18.5 & 34.3 & 11.3 & 26.8 \\[2pt]
LwF\conf{ECCV16} &
  31.9 & 51.8 & 17.6 & 39.2 &
  27.1 & 39.6 & 15.2 & 31.5 \\[2pt]
SDC\conf{CVPR20} &
  40.6 & 56.2 & 32.3 & 46.6 &
  29.5 & 43.8 & 26.3 & 40.6 \\[2pt]
PASS\conf{CVPR21} &
  30.8 & 48.3 & 17.6 & 31.1 &
  24.5 & 39.5 & 18.5 & 30.4 \\[2pt]
FeTrIL\conf{WACV23} &
  34.9 & 51.2 & 23.3 & 37.9 &
  31.0 & 45.3 & 25.9 & 39.9 \\[2pt]
FeCAM\conf{NeurIPS23} &
  32.4 & 48.7 & 21.1 & 34.5 &
  30.9 & 44.9 & 24.9 & 37.9 \\[2pt]
EFC\conf{ICLR24} &
  43.5 & 58.1 & 32.4 & 47.0 &
  34.5 & 47.9 & 28.4 & 42.1 \\[2pt]
ADC\conf{CVPR24} &
  46.5 & 61.4 & 35.1 & 51.7 &
  32.3 & 43.0 & 18.1 & 36.0 \\[2pt]
LDC\conf{ECCV24} &
  45.4 & 59.5 & 35.5 & 51.9 &
  34.2 & 46.8 & 24.9 & 38.2 \\[2pt]
AdaGauss\conf{NeurIPS24} &
  46.8 & 60.9 & 37.9 & 54.4 &
  32.9 & 45.8 & 27.5 & 39.5 \\[2pt]
DPCR\conf{ICML2025} &
  50.2 & 62.8 & 39.8 & 54.8 &
  34.3 & 46.9 & 25.6 & 39.3 \\[2pt]
\textbf{Ours} &
  \textbf{52.0} & \textbf{64.4} & \textbf{43.0} & \textbf{56.6} &
  \textbf{37.2} & \textbf{50.6} & \textbf{31.5} & \textbf{45.1} \\
\bottomrule
\end{tabular*}
\end{table}

\subsection{Main Results}
Tables~\ref{tab:main_results} and \ref{tab:main_results2} present the mean accuracy of our framework (Ours) alongside established baselines. Standard deviations across five runs are provided in \textsc{Supplementary Material}. Across the from-scratch benchmarks, our method yields consistent improvements over the evaluated competitors.

On \textbf{CIFAR-100}, our method achieves competitive performance across both settings. At $T{=}10$, it improves upon DPCR by $+1.8$ pp in $A_{\text{last}}$ and $+1.6$ pp in $A_{\text{inc}}$, and exceeds AdaGauss by $+5.2$ pp and $+3.5$ pp, respectively. A similar trend of final-task retention is observed at $T{=}20$, where it leads DPCR by $+3.2$ pp in $A_{\text{last}}$.

On \textbf{TinyImageNet}, similar performance gains are observed. Our method improves upon DPCR by $+2.9/+3.7$ pp at $T{=}10$ (for $A_{\text{last}}/A_{\text{inc}}$). At $T{=}20$, it exceeds EFC by $+3.1/+3.0$ pp, suggesting that the geometry-aware transport is effective at managing representation drift over longer task sequences.

On \textbf{ImageNet-100} (from scratch), our framework maintains consistent performance. It achieves the highest final-task accuracy ($A_{\text{last}}$) at both $T{=}10$ ($53.2\%$) and $T{=}20$ ($44.7\%$), outperforming baselines like AdaGauss and EFC. While LDC reports a higher running average ($A_{\text{inc}}$) at $T{=}10$, our approach remains competitive incrementally while demonstrating higher final knowledge retention.

Finally, on \textbf{CUB-200} with a pretrained backbone, representation drift is typically milder, leaving less headroom for cross-task alignment. Our method performs comparably to AdaGauss at $T{=}10$, while EFC demonstrates the strongest performance at $T{=}20$ ($46.1/59.3$). Note that DPCR results are omitted (marked as ``--'') because it did not
complete reliably under the pretrained CUB-200 protocol, as discussed in the
reproduction note above. A detailed discussion of the overall CUB-200 results and the effects of pretrained backbones is provided in the \textsc{Supplementary Material}. Overall, these evaluations validate the efficacy of the proposed framework in managing representation drift across diverse incremental learning scenarios.

\begin{table}[t]
\caption{\textbf{ImageNet-100 (from scratch) \& CUB-200 (pretrained).}
Average incremental ($A_{\text{inc}}$) and last-task average ($A_{\text{last}}$) accuracy (\%). \textsuperscript{\dag}: results reported by~\cite{GomezVillaGWBTW24}. Results are averaged over five runs (standard deviations are provided in \textsc{Supplementary Material}). Best in \textbf{bold}.}
\label{tab:main_results2}
\centering
\small
\begin{tabular*}{\linewidth}{@{\extracolsep{\fill}}lcccccccc}
\toprule
\multirow{2}{*}{\textbf{Method}} &
\multicolumn{4}{c}{\textbf{ImageNet-100}} &
\multicolumn{4}{c}{\textbf{CUB-200}} \\
\cmidrule(lr){2-5}\cmidrule(lr){6-9}
& \multicolumn{2}{c}{$T{=}10$} & \multicolumn{2}{c}{$T{=}20$} &
  \multicolumn{2}{c}{$T{=}10$} & \multicolumn{2}{c}{$T{=}20$} \\
\cmidrule(lr){2-3}\cmidrule(lr){4-5}\cmidrule(lr){6-7}\cmidrule(lr){8-9}
& $A_{\text{last}}$ & $A_{\text{inc}}$ &
  $A_{\text{last}}$ & $A_{\text{inc}}$ &
  $A_{\text{last}}$ & $A_{\text{inc}}$ &
  $A_{\text{last}}$ & $A_{\text{inc}}$ \\
\midrule
EWC &
  25.1 & 40.6 & 13.7 & 29.2 &
  15.8 & 32.6 & 12.3 & 27.2 \\[2pt]
LwF\conf{ECCV16} &
  33.4 & 51.5 & 18.6 & 41.3 &
  30.4 & 46.1 & 19.4 & 34.7 \\[2pt]
SDC\conf{CVPR20} &
  35.4 & 50.1 & 19.4 & 36.5 &
  50.3 & 60.5 & 27.9 & 40.1 \\[2pt]
PASS\conf{CVPR21} &
  26.4 & 45.7 & 14.4 & 31.7 &
  27.0 & 42.3 & 18.1 & 36.9 \\[2pt]
FeTrIL\conf{WACV23} &
  36.2 & 52.6 & 26.6 & 42.4 &
  36.9 & 48.2 & 34.6 & 45.3 \\[2pt]
FeCAM\conf{NeurIPS23} &
  38.7 & 54.8 & 29.0 & 44.6 &
  40.2 & 54.9 & 36.2 & 48.9 \\[2pt]
EFC\conf{ICLR24} &
  50.9 & 61.3 & 38.6 & 50.5 &
  51.0 & 63.3 & \textbf{46.1} & \textbf{59.3} \\[2pt]
ADC\conf{CVPR24} &
  38.3 & 55.5 & 25.1 & 43.4 &
  49.5 & 58.8 & 35.4 & 48.3 \\[2pt]
LDC\conf{ECCV24} &
  51.4\textsuperscript{\dag} & \textbf{69.4}\textsuperscript{\dag} & 28.5 & 46.5 &
  47.5 & 55.7 & 27.2 & 39.8 \\[2pt]
AdaGauss\conf{NeurIPS24} &
  51.1 & 65.0 & 42.6 & 57.4 &
  \textbf{52.9} & \textbf{63.4} & 45.0 & 57.0 \\[2pt]
DPCR\conf{ICML2025} &
  49.9 & 64.8 & 37.3 & 54.7 &
  -- & -- & -- & -- \\[2pt]
\textbf{Ours} &
  \textbf{53.2} & 67.4 & \textbf{44.7} & \textbf{59.1} &
  52.4 & 63.1 & 43.7 & 55.8 \\
\bottomrule
\end{tabular*}
\end{table}

\subsection{Component Study: Validating the Geometry-Anchored Framework}
\label{sec:ablation_gls_mechanism}

Our framework integrates an Analytic Geometric Anchor (AGA) into a Topology-Aware Evolution process. Since the AGA operates as an explicit closed-form prior, it relies on the topology-aware training phase to function effectively. We sequentially ablate these components in Table~\ref{tab:abl_components} to evaluate their respective contributions.

\paragraph{\textbf{Impact of the Analytic Geometric Anchor.}}
The baseline (\xmark~Top-Aware, \xmark~AGA, \cmark~Refine) structurally aligns with the AdaGauss paradigm: open-loop optimization followed by a decoupled adapter. Upgrading this to include topology-aware constraints (\cmark~Top-Aware, \xmark~AGA, \cmark~Refine) allows the adapter to be learned during the primary training phase. However, without the AGA prior, the adapter effectively functions as a standard MLP mapping $z_{\mathrm{old}} \to z_{\mathrm{new}}$. While this restricts manifold distortion—improving CIFAR-100 $A_{\text{last}}$ from $46.8\%$ to $50.3\%$—relying solely on this neural projection is limited. Such a mapping optimizes a standard Euclidean objective, which may not fully account for the Mahalanobis-sensitive tail errors that impact Bayes evaluation.

\paragraph{\textbf{Comparing end-to-end evolution and post-hoc refinement.}}
Performance improves further when the AGA is embedded as a geometric prior, transforming the adapter's role from a direct projection to a localized residual correction. Notably, our results suggest that decoupled post-hoc fine-tuning does not necessarily improve well-calibrated representations. As shown in Table~\ref{tab:abl_components}, the end-to-end variant without secondary refinement (gray row: \xmark~Refine) generally performs better than the version with refinement (\cmark~Refine) across most settings, reaching $52.0\%$ on CIFAR-100. 

This indicates that the geometry-anchored framework reaches a stable configuration during the active training phase. Isolated fine-tuning of the adapter after freezing the backbone may disrupt the coupling established during co-evolution, potentially causing the adapter to overfit to static features. An exception is observed in the longer task sequence of TinyImageNet ($T{=}20$), where accumulated drift makes isolated calibration marginally beneficial ($31.7\%$ vs $31.5\%$). Overall, these results indicate that embedding transport as a primary training constraint provides a robust one-stage framework, reducing the need for retroactive fine-tuning.

\begin{table}[t]
  \caption{\textbf{Component ablation} of the Geometry-Anchored Transport Framework on CIFAR-100 and TinyImageNet ($T{=}10$ and $T{=}20$). We ablate the Topology-Aware Evolution (\textbf{Top-Aware}), the Analytic Geometric Anchor (\textbf{AGA}), and the optional post-hoc refinement (\textbf{Refine}). Best results in \textbf{bold}.}
  \label{tab:abl_components}
  \centering
  \small
  \setlength{\tabcolsep}{3pt}
  \renewcommand{\arraystretch}{1.0}
  \resizebox{\linewidth}{!}{%
  \begin{tabular}{@{}ccc|cccc|cccc@{}}
    \toprule
    \multicolumn{3}{c|}{\textbf{Components}} &
    \multicolumn{4}{c|}{\textbf{CIFAR-100}} &
    \multicolumn{4}{c}{\textbf{TinyImageNet}} \\
    \cmidrule(lr){4-7}\cmidrule(lr){8-11}
    \multicolumn{3}{c|}{} &
    \multicolumn{2}{c}{\textit{$T{=}10$}} &
    \multicolumn{2}{c|}{\textit{$T{=}20$}} &
    \multicolumn{2}{c}{\textit{$T{=}10$}} &
    \multicolumn{2}{c}{\textit{$T{=}20$}} \\
    \cmidrule(lr){1-3}
    \cmidrule(lr){4-5}\cmidrule(lr){6-7}
    \cmidrule(lr){8-9}\cmidrule(lr){10-11}
    Top-Aware &
    AGA &
    Refine &
    $A_{\text{last}}$ & $A_{\text{inc}}$ &
    $A_{\text{last}}$ & $A_{\text{inc}}$ &
    $A_{\text{last}}$ & $A_{\text{inc}}$ &
    $A_{\text{last}}$ & $A_{\text{inc}}$ \\
    \midrule
    % 1. Baseline (AdaGauss) - Open-loop + Post-hoc adapter only
    \xmark & \xmark & \cmark &
      46.8 & 60.9 & 37.9 & 54.4 & 32.9 & 45.8 & 27.5 & 39.5 \\
      
    % 2. Top-Aware Only - Symbiotic grounding + Neural adapter (Euclidean loss misses Mahalanobis tail)
    \cmark & \xmark & \cmark &
      50.3 & 63.1 & 41.6 & 55.9 & 35.6 & 49.3 & 30.4 & 44.2 \\
      
    % 3. Ours (Full w/ Refine) - Shows traditional two-stage
    \cmark & \cmark & \cmark &
      51.7 & 64.3 & 42.3 & 56.3 & 37.0 & 50.4 & \textbf{31.7} & \textbf{45.3} \\
      
    \midrule
    % 4. Ours (End-to-End only) - The ultimate performance (No Refine needed)
    \rowcolor{gray!15}
    \cmark & \cmark & \xmark &
      \textbf{52.0} & \textbf{64.4} & \textbf{43.0} & \textbf{56.6} & \textbf{37.2} & \textbf{50.6} & 31.5 & 45.1 \\
    \bottomrule
  \end{tabular}%
  }
\vspace{-2mm}
\end{table}

\begin{figure*}[t]
  \centering
  \begin{subfigure}[t]{0.31\linewidth}
    \centering
    \includegraphics[width=\linewidth]{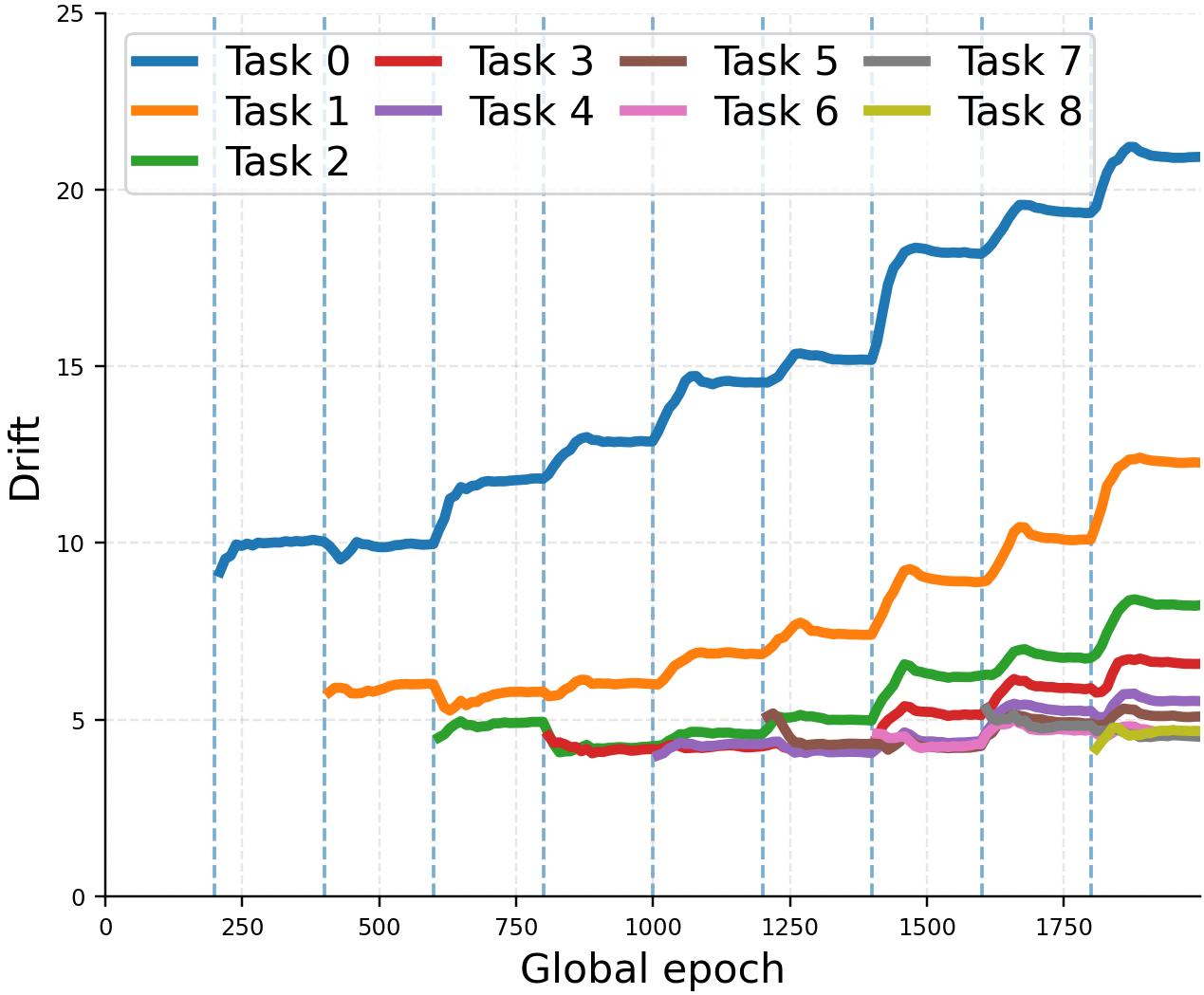}
    \caption{AdaGauss}
    \label{fig:drift-dwkd}
  \end{subfigure}
  \hfill
  \begin{subfigure}[t]{0.31\linewidth}
    \centering
    \includegraphics[width=\linewidth]{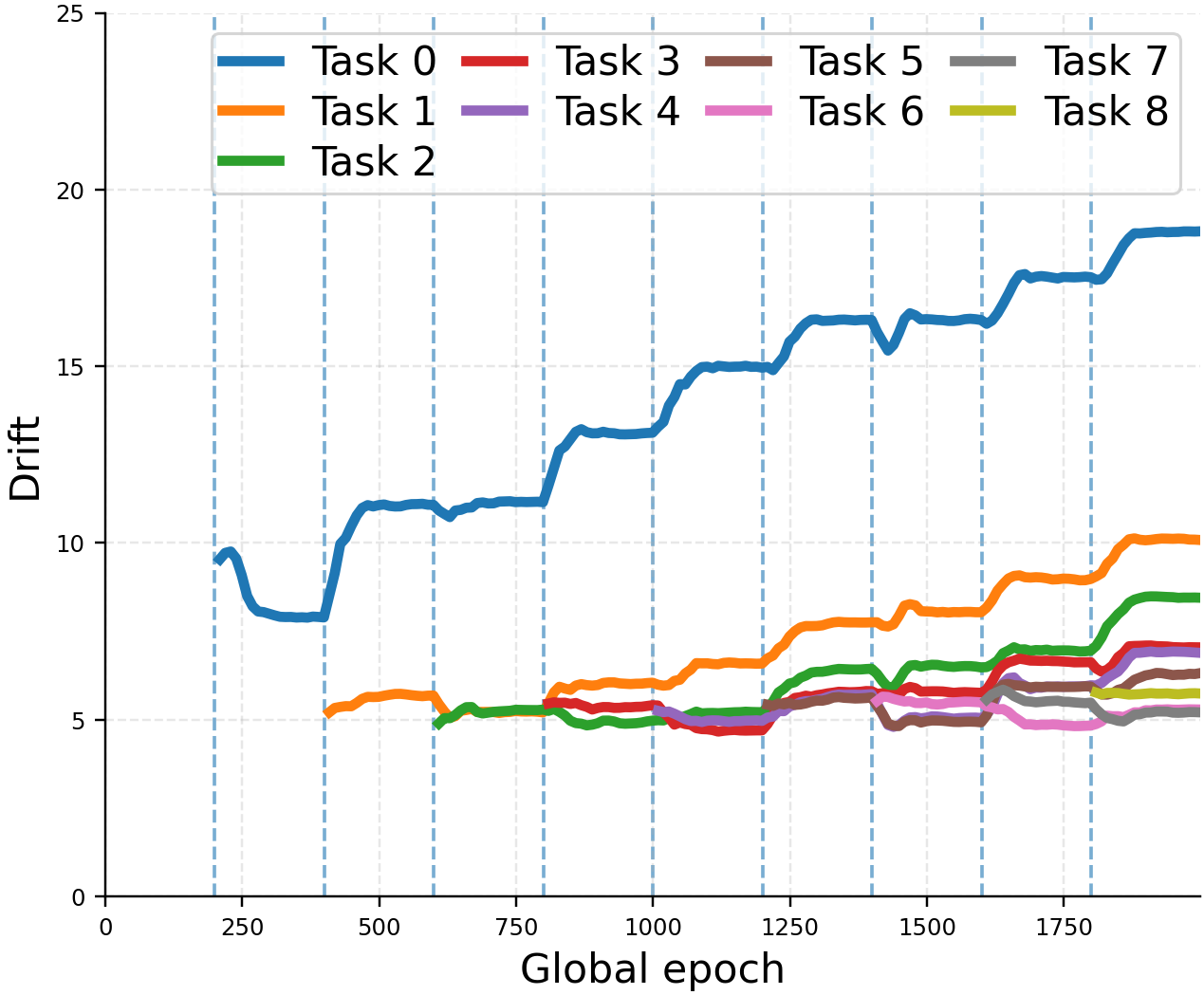}
    \caption{Ours (w/o AGA)}
    \label{fig:drift-no-dwkd}
  \end{subfigure}
  \hfill
  \begin{subfigure}[t]{0.31\linewidth}
    \centering
    \includegraphics[width=\linewidth]{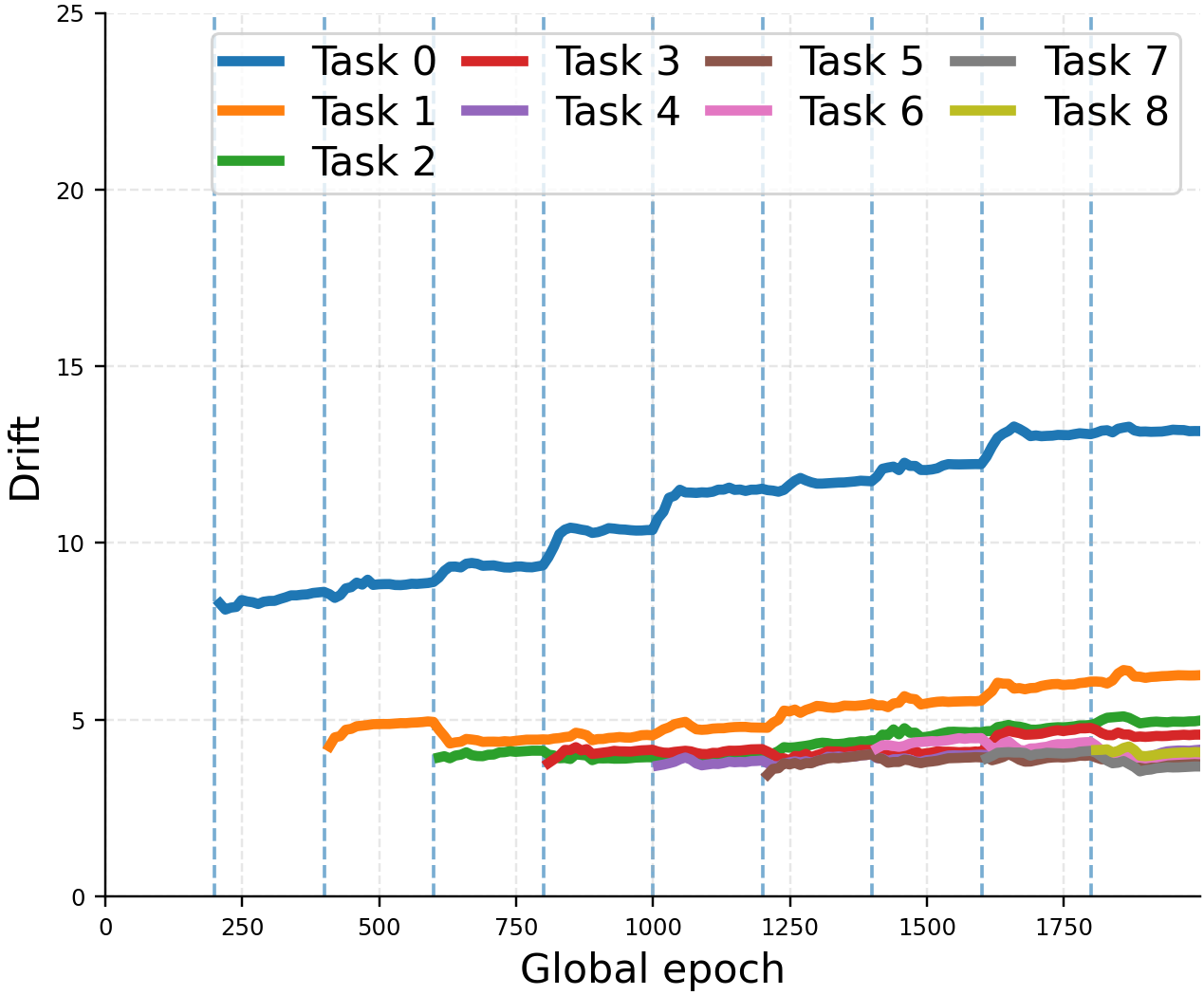}
    \caption{Ours (Full)}
    \label{fig:drift-final}
  \end{subfigure}
  \caption{\textbf{Prototype drift over global epochs} on Tiny-ImageNet (10 steps).
We track, for each past task $k$, the mean feature $\mu_c^{(e)}$ of class $c$ under the current backbone at global epoch $e$ against its stored/transported prototype $\hat{\mu}_c^{(e)}$ used for Bayes evaluation, reporting
$\mathrm{Drift}_k(e)=\frac{1}{|\mathcal{C}_k|}\sum_{c\in\mathcal{C}_k}\|\mu_c^{(e)}-\hat{\mu}_c^{(e)}\|_2$.
The topology-aware anchoring alone (b) reduces drift relative to AdaGauss (a); calibrating the residual to the explicit AGA prior (c) further suppresses drift, particularly limiting error accumulation for earlier tasks.}
  \label{fig:drift-ablations}
\end{figure*}

\begin{figure}[t]
  \centering
  % 1. Mahalanobis Tail 
  \begin{subfigure}[b]{0.32\linewidth}
    \centering
    \includegraphics[width=\linewidth,trim=2pt 2pt 2pt 2pt,clip]{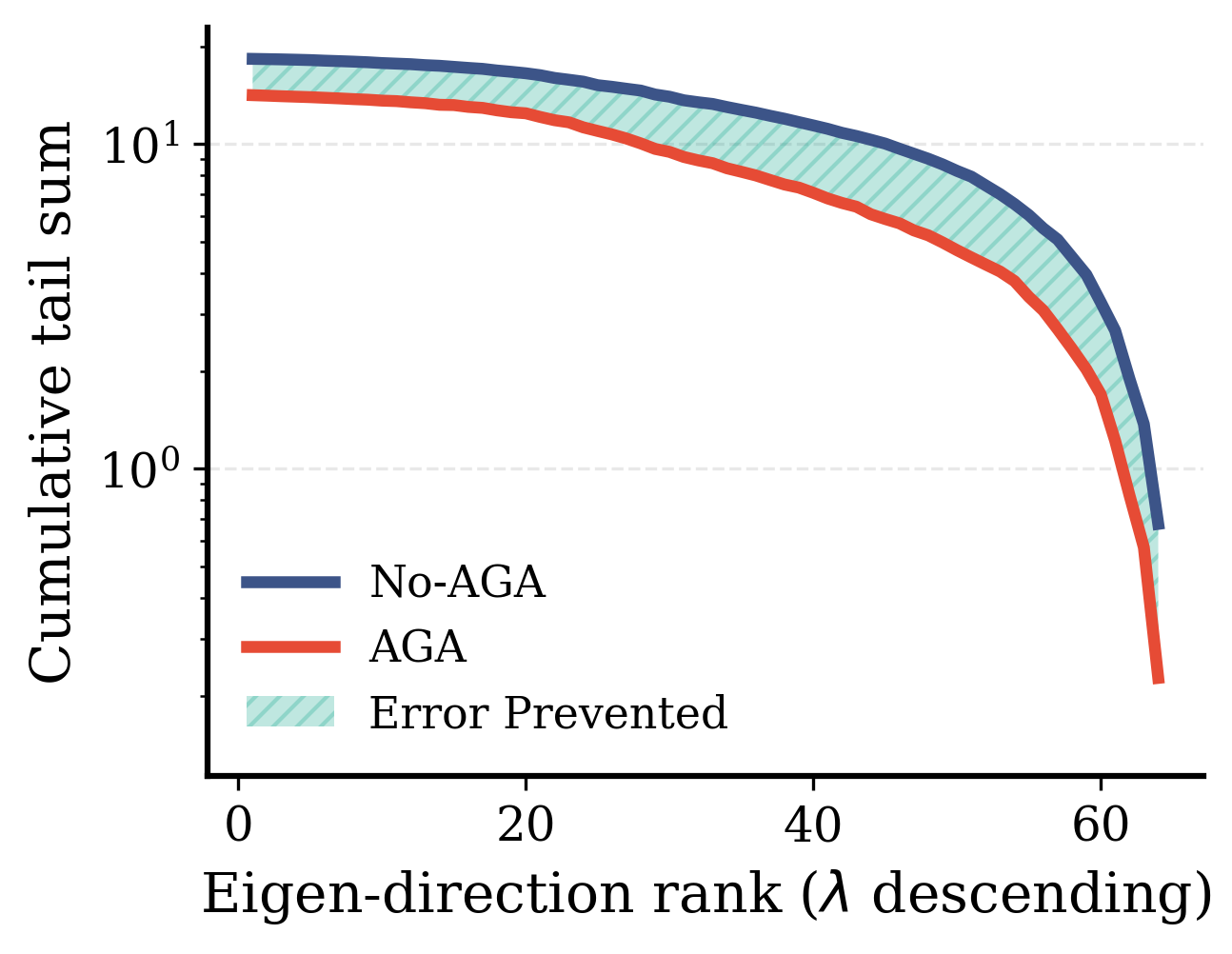}
    \caption{Maha-tail (Task 0)}
    \label{fig:ablation_maha}
  \end{subfigure}\hfill
  % 2. Margin Density
  \begin{subfigure}[b]{0.32\linewidth}
    \centering
    \includegraphics[width=\linewidth,trim=2pt 2pt 2pt 2pt,clip]{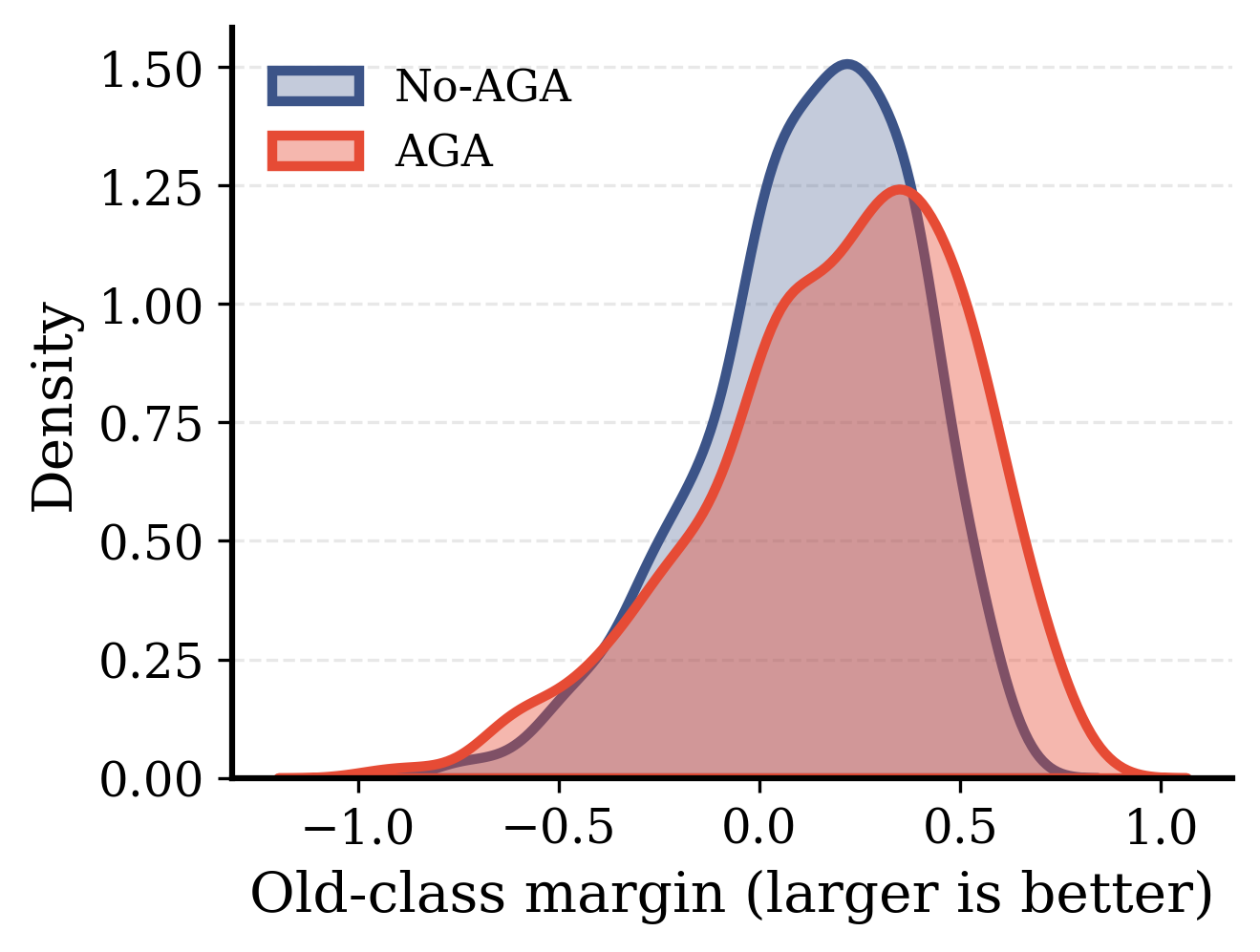}
    \caption{Margin density (Task 0)}
    \label{fig:ablation_margin}
  \end{subfigure}\hfill
  % 3. Old->New Confusion
  \begin{subfigure}[b]{0.32\linewidth}
    \centering
    \includegraphics[width=\linewidth,trim=2pt 2pt 2pt 2pt,clip]{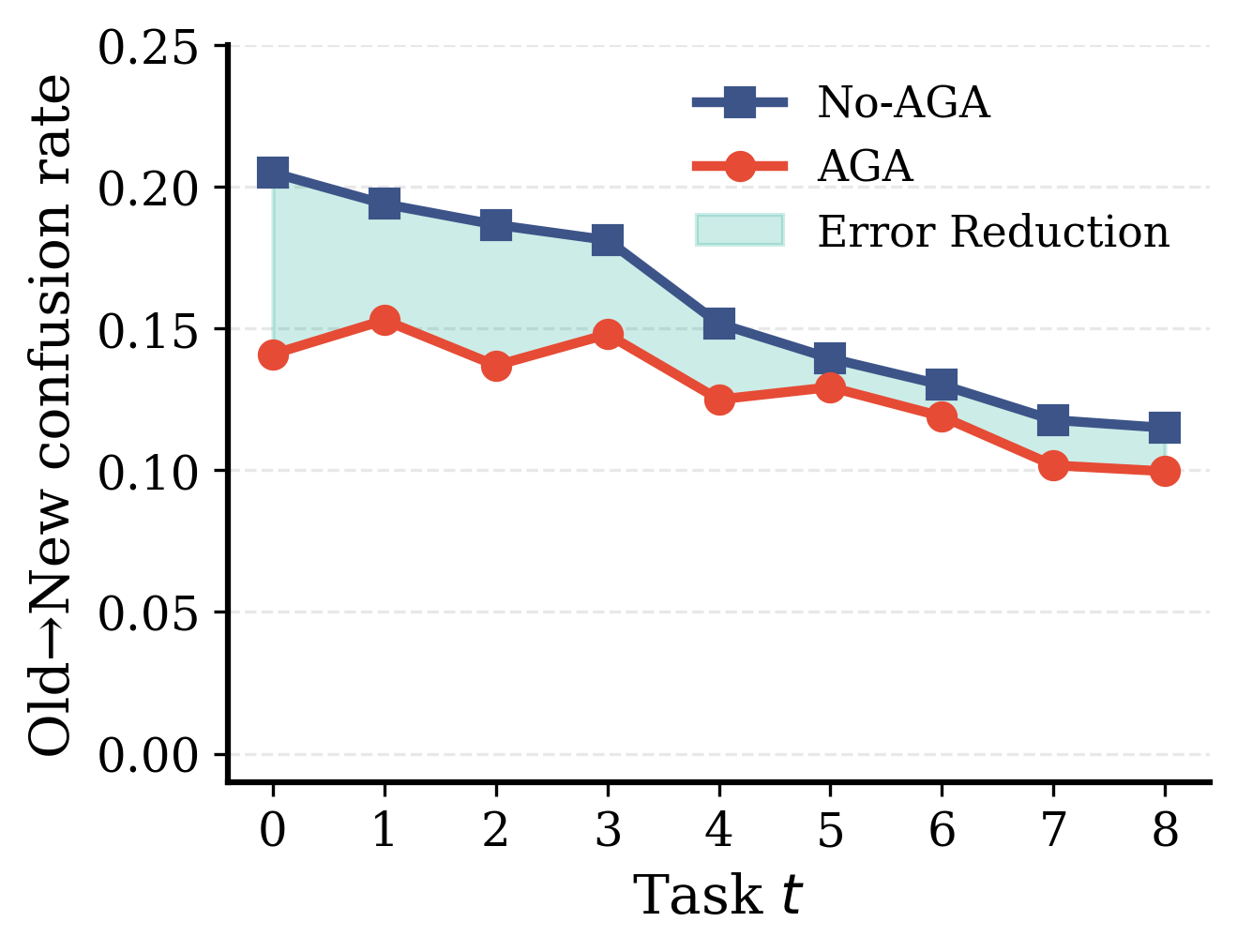}
    \caption{Old$\to$New rate}
    \label{fig:ablation_old2new}
  \end{subfigure}
  
  \vspace{-2mm} 
  
  \caption{\textbf{Diagnostic ablations for the Analytic Geometric Anchor (AGA).} 
  \textbf{(a)} Cumulative Mahalanobis tail $T_t$; the shaded region highlights the tail error mitigated by the AGA. 
  \textbf{(b)} Estimated probability density of old-class margins; integrating the AGA shifts the distribution toward positive values. 
  \textbf{(c)} Old$\to$New confusion rate $\rho_t$ across tasks; the geometric prior reduces misclassifications of legacy samples. Together, these diagnostics suggest that targeting Mahalanobis-sensitive tail error translates into more stable decision boundaries.}
  \label{fig:ablation_main}
  \vspace{-3mm} 
\end{figure}

\paragraph{\textbf{Our framework reduces prototype drift.}}
We first evaluate the preservation of stored class prototypes. Fig.~\ref{fig:drift-ablations} tracks prototype drift over global epochs by comparing the current class-mean feature to the corresponding transported prototype used for Bayes evaluation. While the anchored baseline variant (Ours w/o AGA) limits initial topological changes compared to AdaGauss, enforcing the explicit AGA prior (Ours) further suppresses this drift. This stabilization is noticeable for earlier tasks, where representation changes accumulate over time, indicating that the analytic anchor helps maintain faithfulness in the old$\!\to$new transport.

\paragraph{\textbf{Suppressing Mahalanobis-sensitive tail error (Fig.~\ref{fig:ablation_main}a).}}
We report the cumulative Mahalanobis tail $T_t$, which characterizes the drift sensitivity across the feature spectrum (formulated in the \textsc{Supplementary Material}). The Mahalanobis metric assigns larger weights to feature directions with naturally small variance (the spectrum's ``tail''). Consequently, parameter drifts along these directions are amplified, contributing to forgetting. The transport curve for our full framework (Ours) bounds the non-AGA baseline (Ours w/o AGA) from below. The shaded region reveals that the magnitude of error mitigated by the AGA increases at high eigen-ranks (the tail where $\lambda_i \to 0$). This observation supports the anisotropic shrinkage established in Theorem~\ref{theorem1}: governed by the spectral denominator $\gamma_j + \rho\,\lambda_i$, the closed-form prior dampens updates along sensitive dimensions, mitigating the error components that affect Mahalanobis evaluation.

\paragraph{\textbf{Preserving old-class decision margins (Fig.~\ref{fig:ablation_main}b).}}
We estimate the old-class margin $m_t(x)$, representing the log-likelihood gap between the ground-truth and the nearest competing class (refer to the \textsc{Supplementary} for the precise definition). A larger positive margin indicates higher robustness against new-class interference. As shown, the geometry-anchored framework shifts the density peak toward larger, positive values compared to the baseline. This separation between the legacy class and competitors aligns with the tail-error mitigation observed in Fig.~\ref{fig:ablation_main}a. These findings support our theoretical premise (Theorem~\ref{theorem2}): targeting Mahalanobis-sensitive tail error helps stabilize decision boundaries.

\paragraph{\textbf{Mitigating old$\to$new confusions (Fig.~\ref{fig:ablation_main}c).}}
We track the Old$\to$New confusion rate $\rho_t$, which measures the empirical probability of misclassifying legacy samples into newly introduced classes (detailed in the \textsc{Supplementary Material}). The addition of the AGA consistently reduces $\rho_t$ across the continual learning trajectory relative to the non-AGA baseline. The shaded area illustrates the framework's capability in limiting legacy samples from being misclassified into new classes. These improvements are consistent with the tail suppression and margin preservation established in earlier diagnostics.

\paragraph{\textbf{Overhead of online prior estimation.}}
Our framework periodically refreshes the transport prior during training. For $E$ training epochs per task and a refresh interval of $K$ epochs, we perform $R \approx E/K$ refreshes. Each refresh samples $B_t$ mini-batches, requiring two additional forward passes per batch (one each for the frozen and current backbones). To maintain consistent relative compute across varying task partitions, we scale $B_t$ proportionally to the optimization steps per epoch $S_t$ (i.e., $B_t=\gamma S_t$). The relative computational overhead therefore scales as:
\[
\frac{2 R B_t}{c E S_t}
\;\approx\;
\frac{2\gamma}{cK},
\]
rendering it relatively independent of the specific task partition. Under our default configuration ($E{=}200, K{=}10$), and taking one refresh to sample approximately one epoch's worth of steps ($\gamma{=}1$) with a standard approximation that one training step costs about three forward passes ($c{\approx}3$), the relative overhead is:
\[
\frac{2\gamma}{cK}
=
\frac{2}{3 \cdot 10}
=
\frac{1}{15}
\approx 6.7\%.
\]
This corresponds to a moderate single-digit percentage overhead. Furthermore, because the GLS closed-form solution (Theorem~\ref{theorem1}) operates exclusively on low-dimensional feature covariances rather than high-dimensional image tensors, the matrix solver overhead is virtually negligible. These factors confirm that analytic anchoring remains highly computationally efficient in practice.

\section{Conclusion, Limitations, and Future Work}
\label{sec:conclusion}

\paragraph{\textbf{Conclusion.}} We introduced the Geometry-Anchored Transport Framework, formulating feature transport in exemplar-free class-incremental learning as an integrated constraint during representation learning rather than a strictly post-hoc procedure. Our method combines an Analytic Geometric Anchor with Topology-Aware Evolution to regularize the legacy manifold, while a residual network calibrates the non-linear deformations. This coupled design facilitates the transport of cached Gaussian statistics for classification, addressing the sensitivity of decision margins to moment-transport errors.

\paragraph{\textbf{Limitations.}} Our approach is currently designed for prototype preservation with Bayes-style evaluation, and its theoretical guarantees are aligned with this parametric setting. The analytic anchor relies on an affine prior with EMA updates; while the residual network accommodates local non-linearities, severe non-linear or highly non-stationary representation shifts may still challenge this linearized global approximation, particularly when covariance estimates are noisy or poorly conditioned. 

\paragraph{\textbf{Future work.}} Future research could explore generalizing geometry-anchored transport beyond single class-conditional Gaussians (e.g., using Gaussian mixtures~\cite{KoryckiK24} or implicit density surrogates~\cite{ZhangZZS23}) while maintaining margin stability bounds. Additionally, investigating alternative anchor formulations (e.g., low-rank, kernelized, or orthogonality-constrained priors~\cite{KohSBSHP0C23}) may offer ways to model complex global deformations while preserving the separation of analytic priors and neural residuals.

\paragraph{\textbf{\textsc{Supplementary Material.}}} Due to space constraints, extended details are deferred to the supplementary material. Key contents include: (1) complete mathematical proofs for Theorems~\ref{theorem1} and~\ref{theorem2}; (2) a unified formulation of prior post-hoc feature transport methods; (3) algorithmic pseudocode, architectural specifications (e.g., feature dimensions), and an extended analysis of MLP computational overhead; and (4) comprehensive experimental configurations, encompassing full hyperparameter settings, sensitivity analyses, formal diagnostic metric definitions, and our rationale for baseline (AdaGauss) selection. Additionally, the supplementary material reports classifier-robustness results with an alternative sampling-based linear head, showing that the gains persist beyond the default Mahalanobis readout.

\bibliographystyle{splncs04}
\bibliography{main}

\end{document}